\documentclass[conference]{IEEEtran}
\IEEEoverridecommandlockouts

\usepackage{cite}
\usepackage{amsmath,amssymb,amsfonts}
\usepackage{algorithmic}
\usepackage{graphicx}
\usepackage{textcomp}
\usepackage{hyperref}
\usepackage{xcolor}
\usepackage{booktabs}
\usepackage{multirow}
\def\BibTeX{{\rm B\kern-.05em{\sc i\kern-.025em b}\kern-.08em
    T\kern-.1667em\lower.7ex\hbox{E}\kern-.125emX}}

\title{Generalize Your Face Forgery Detectors: \protect\\ An Insertable Adaptation Module Is All You Need}

\author{\IEEEauthorblockN{ Xiaotian Si$^1$, Linghui Li$^1$$^\star$, Liwei Zhang$^1$, Ziduo Guo$^1$, Kaiguo Yuan$^1$, Bingyu Li$^2$, Xiaoyong Li$^1$\thanks{$^\star$Corresponding author}}
\IEEEauthorblockA{$^1$Key Laboratory of Trustworthy Distributed Computing and Service (MoE),\\
Beijing University of Posts and Telecommunications, Beijing, China} 
\IEEEauthorblockA{$^2$School of Cyber Science and Technology, Beihang University, Beijing, China}
\IEEEauthorblockA{\{sixt,lilinghui,lw\_z,DUODUOBOOM,flyingdreaming,lixiaoyong\}@bupt.edu.cn, libingyu@buaa.edu.cn}
}

\usepackage[switch]{lineno}
\begin{document}

\maketitle

\begin{abstract}
A plethora of face forgery detectors exist to tackle facial deepfake risks. However, their practical application is hindered by the challenge of generalizing to forgeries unseen during the training stage.
%
To this end, we introduce an insertable adaptation module that can adapt a trained off-the-shelf detector using only online unlabeled test data, without requiring modifications to the architecture or training process.
Specifically, we first present a learnable class prototype-based classifier that generates predictions from the revised features and prototypes, enabling effective handling of various forgery clues and domain gaps during online testing. Additionally, we propose a nearest feature calibrator to further improve prediction accuracy and reduce the impact of noisy pseudo-labels during self-training.
%
Experiments across multiple datasets show that our module achieves superior generalization compared to state-of-the-art methods. 
Moreover, it functions as a plug-and-play component that can be combined with various detectors to enhance the overall performance.
\end{abstract}


\section{Introduction}
\label{sec:intro}
Facial deepfakes have gained significant popularity on the internet and social media platforms, driven by rapid advancements in forgery techniques~\cite{xu2022high,sun2022fenerf,hsu2022dual}. Despite deepfakes finding applications in entertainment, particularly in film production, they also pose risks, including the potential to damage the reputation of public figures and accelerate the spread of misinformation.
To counteract these negative impacts, face forgery detection have been developed to automatically distinguish between genuine and forgery facial images, attracting increasing attention within the research community~\cite{chen2021local,wang2021representative,liu2021spatial,yan2023ucf,yan2024transcending}.

Most existing methods treat face forgery detection as a binary classification task~\cite{li2020face,gu2022exploiting,haliassos2022leveraging,shiohara2022detecting,yan2023ucf,yan2024transcending}, typically employing deep neural networks trained on datasets with both authentic and forged faces. Although these methods perform well on datasets that are similar to their training data, their effectiveness significantly diminishes in cross-dataset scenarios, particularly when facing novel forgeries that deviate from the training distribution. 
To tackle this challenge and achieve domain generalization, one common type of strategy focuses on generalizing detectors during the training phase. Related approaches include data augmentation-based methods~\cite{li2020face,shiohara2022detecting,yan2024transcending} that generate pseudo forgeries, compelling detectors to focus on blending artifacts, multi-task learning strategies~\cite{yan2023ucf,hu2021improving} that aim to extract common forgery features for improved generalization, and frequency artifacts~\cite{qian2020thinking,liu2021spatial} that exploit frequency information in addition to the RGB domain. 
While these methods can mitigate generalization issues to some extent, overcoming biases in target test data within practical testing environments remains challenging.

Another type of strategy aims to generalize the detector to new target data during evaluation by leveraging domain adaptation methods. For instance, OST~\cite{chen2022ost} introduces a test-time adaptation framework that constructs a test sample-specific auxiliary task to update the model before applying it to the sample.
Despite its success, this adaptation approach is closely tied to the meta-training procedure of the face forgery detector, making it ineffective for off-the-shelf detectors with various training processes, which is often a practical requirement.
To achieve this, we aim to introduce an insertable adaptation module that is versatile across various detectors and can effectively address generalization challenges in online testing scenarios. This module should be designed for seamless integration, without altering the architecture or training process of the detectors. 

While there are state-of-the-art test-time adaptation methods~\cite{lee2013pseudo,iwasawa2021test,jang2023test} that meet the aforementioned criteria, our experiments reveal that they suffer from instability and decreased performance in the context of face forgery detection due to the large domain gaps created by the diverse forgery clues present in different datasets.
\begin{figure*}[!t] 
\centering 
\includegraphics[width=0.95\textwidth]{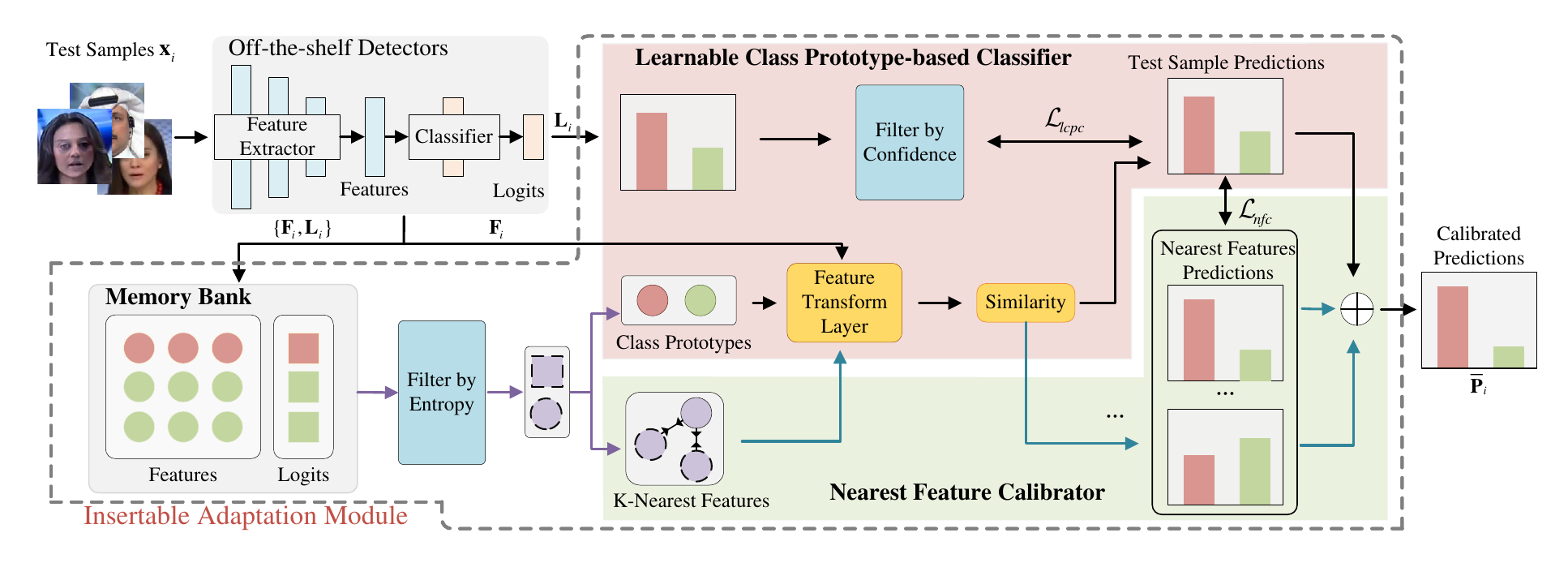} 
\caption{
Overview of our proposed adaption module for the off-the-shelf face forgery detectors. 
} 
\label{fig:framework} 
\end{figure*}
To overcome these challenges, we introduce a learnable class prototype-based classifier, that establishes a trainable pathway from the original features to generate the adapted features and prototypes for predicting the test data, effectively handling various forgery patterns. The parameters of the classifier are updated through a self-training procedure during testing. To further mitigate the impact of noisy pseudo-labels and enhance final prediction accuracy, we propose a nearest feature calibrator. This calibrator makes predictions collectively using the test sample and its nearest samples in the feature space, with consistency regularization that encourages similar predictions for the test sample and its nearest features during test-time adaptation.




\section{METHODOLOGY}
\label{sec:Approach}
Fig.~\ref{fig:framework} illustrates the overall pipeline of our face forgery detection method. Our approach utilizes an off-the-shelf detector trained on the source domain $\mathbb{D}_{s}$ as the base detector, coupled with our insertable adaptation module to improve performance on unlabeled test samples $\mathbf{x}_{i} \in \mathbb{D}_{t}$ from the target domain $\mathbb{D}_{t}$, which contains new forgeries not present in $\mathbb{D}_{s}$. This adaptation module is insertable and processes only the output from the base detector without modifying its architecture or training process.
The module includes a memory bank~\ref{ssec:Overview}) to retain the historical information during evaluation, a learnable class prototype-based classifier (ref to sec.~\ref{ssec:Learnable Class Prototypes-based Classifier}) designed to handle diverse forgery clues and enhance the domain adaptation flexibility, and a nearest feature calibrator (ref to sec.~\ref{ssec:Nearest Feature Calibrator}) proposed to further enhance prediction accuracy and mitigate the impact of noisy labels. 

The predictions of the test samples $\mathbf{x}_{i}$ are generated using Eq.~\ref{eq7}, and the parameters in the adaptation module are updated over a few steps using Eq.~\ref{eq9}. It is worth to noting that the base detector remains fixed to ensure stable adaptation.

\subsection{Memory Bank}
\label{ssec:Overview}
The adaptation module maintains a memory bank to store the historical information during online testing. Formally, the base detector ${\rm g}$ is composed of a feature extractor ${\rm f}$ and a classifier ${\rm h}$, such that ${\rm g}={\rm f} \circ {\rm h}$. Given a batch of unlabeled test samples $\mathbf{x}_{i}$, we first extract features $\mathbf{F}_{i} = {\rm f}(\mathbf{x}_{i}) \in \mathbb{R}^d$ and then compute the logits $\mathbf{L}_{i} = {\rm h}(\mathbf{F}_{i}) \in \mathbb{R}^c$. 

The memory bank $\mathbb{M} = \{(\mathbf{F},\mathbf{L})\}$ is initialized with the weights of the classifier ${\rm h}$, serving as the initial class prototypes, and is updated with newly arrived test samples: $\mathbb{M} = \mathbb{M} \cup (\mathbf{F}_{i},\mathbf{L}_{i})$. It also filters out unreliable elements with high entropy: ${\rm H}(\mathbf{L}_{i}) = -\sum \sigma(\mathbf{L}_{i}){\rm log}(\sigma(\mathbf{L}_{i}))$, where $\sigma$ is the softmax function. For each predicted class $\hat{\mathbf{L}}_{i} = {\rm argmax} (\mathbf{L}_{i})$, elements with the top-$K$ entropy will be filtered out, and the memory bank is limited to a maximum size of $N_m$.

\subsection{Learnable Class Prototype-based Classifier}
\label{ssec:Learnable Class Prototypes-based Classifier}
To handle diverse forgery clues and achieve stable and reliable adaptation during testing, a learnable class prototype-based classifier is introduced. For generating predictions for test samples $\mathbf{x}_{i}$, we first create prototypes for each class $k$:
\begin{equation}
  \mathbf{C}_{k} = \frac{\sum_j \mathbb{I}(\hat{\mathbf{L}}_{j} = k) \mathbf{F}_{j}}{\sum_j \mathbb{I}(\hat{\mathbf{L}}_{j} = k)},
\end{equation}
where $\mathbb{I}(\cdot)$ is an indicator function, and $j \in \{1,...,N_m\}$ represents the index of elements in $\mathbb{M}$.

After that, we apply $N_t$ trainable feature transform layers ${\rm T}_r, r \in \{1,...,N_t\}$ to obtain the revised features $\mathbf{F}_{i}^{r}$ and class prototypes $\mathbf{C}_{k}^{r}$: $\mathbf{F}_{i}^{r} = {\rm T}_r(\mathbf{F}_{i}) \in \mathbb{R}^{d_t}$, $\mathbf{C}_{k}^{r} = {\rm T}_r(\mathbf{C}_{k}) \in \mathbb{R}^{d_t}$. 
Each ${\rm T}_r$ is a fully-connected network. 
The prototype-based predictions will be the softmax over the similarity between the revised features and the prototypes:
\begin{align}
  \mathbf{P}_{i}^{k}(r) &= \frac{{\rm exp}({\rm S}(\mathbf{F}_{i}^r,\mathbf{C}_{k}^r))}{\sum_{k^{'}} {\rm exp}({\rm S}(\mathbf{F}_{i}^r,\mathbf{C}_{k^{'}}^r))}, \\
  \mathbf{P}_{i}^{k} &= \frac{\sum_r \mathbf{P}_{i}^{k}(r)}{N_t},
\end{align}
where ${\rm S}(\cdot)$ is the cosine similarity function, $\mathbf{P}_{i}^{k}(r)$ represents the prediction probability for class $k$ from the $r$-th transform layers, and $\mathbf{P}_{i}^{k}$ is the final prediction averaged across all $N_t$ transform layers.

We then update all feature transform layers ${\rm T}_r$ using a self-training procedure. Batch predictions $\sigma(\mathbf{L}_{i})$ from the base detector guide the final predictions $\mathbf{P}_{i}$ from the transform layers, to mitigate the impact of the noisy labels, we apply a confidence filter to ignore unreliable predictions:
\begin{align}
   \mathcal{L}_i &= {\rm CE}(\sigma(\mathbf{L}_{i}), \mathbf{P}_{i}), \\
   \mathcal{L}_{lcpc} &= \frac{\sum_i \mathcal{L}_i \mathbb{I}({\rm max}(\mathbf{L}_{i}) > Conf)}{\sum_i \mathbb{I}({\rm max}(\mathbf{L}_{i}) > Conf)}, \label{eq4}
\end{align}
where ${\rm CE}$ is the standard cross-entropy loss, $Conf$ is a confidence threshold, and ${\rm max}(\cdot)$ is the maximum operation.

\subsection{Nearest Feature Calibrator}
\label{ssec:Nearest Feature Calibrator}
To further enhance the prediction accuracy and reduce the negative impact of noisy labels from the base detector, we present a nearest feature calibrator. Using the features $\mathbf{F}_{i}$ of test samples $\mathbf{x}_{i}$ and those in the memory bank $ \mathbf{F}_{j} \in \mathbb{M}$, we find $N_{f}$ nearby elements $(\mathbf{F}_j, \mathbf{L}_j)$ of test samples in $\mathbb{M}$:
\begin{equation}
  \mathbb{N}(\mathbf{F}_{i}, \mathbb{M}) = \{ (\mathbf{F}_{j},\mathbf{L}_{j}) \in \mathbb{M} | {\rm dis}(\mathbf{F}_{i}, \mathbf{F}_{j}) \leq \beta(N_{f}) \},
\end{equation}
where ${\rm dis}(\cdot)$ is a distance function, such as euclidean distance or cosine similarity, $\beta(N_{f})$ is the distance between $\mathbf{F}_{i}$ and the $N_{f}$-th nearest features of $\mathbf{F}_{i}$ from $\mathbb{M}$.

We then obtain the prototype predictions for these nearest features: $\mathbf{P}_{i}(n), n \in \{1,...,N_f\}$. To calibrate the predictions $\mathbf{P}_{i}$ for test samples $\mathbf{x}_{i}$, we average the predictions of $\mathbf{x}_{i}$ and its nearest features, resulting in the calibrated predictions:
\begin{equation}
  \overline{\mathbf{P}}_{i} = \frac{\mathbf{P}_{i} + \sum_n \mathbf{P}_{i}(n)}{N_{f}+1}. \label{eq7}
\end{equation}

To further reduce the negative impact of noisy labels from base detector during self-training, we introduce a consistency regularization term that enforces alignment between the predictions of test samples and those of their nearest features:
\begin{equation}
  \mathcal{L}_{nfc} = \sum_{n} {\rm CE}(\mathbf{P}_{i}(n), \mathbf{P}_{i}). \label{eq8}
\end{equation}
The intuition behind this regularization is that the prototype prediction distributions of test samples should be similar to those of their nearest features.

\subsection{Overall Loss Function}
\label{ssec:Overall Loss Function}
During the online evaluation, to better represent the target data, the trainable feature transform layers ${\rm T}_r$ will be updated over $K_s$ steps for each batch of test samples using the overall loss function:
\begin{equation}
  \mathcal{L} = \mathcal{L}_{lcpc} + \alpha \mathcal{L}_{nfc}, \label{eq9}
\end{equation}
where $\alpha$ is the hyperparameter to balance the importance of consistency regularization term.

\section{EXPERIMENTS}
\label{sec:EXPERIMENTS AND RESULTS }

\subsection{Datasets}
\label{ssec:Datasets}
{\bf Dataset.}
To evaluate the performance of our proposed method, we utilize several widely-used face forgery datasets: FaceForensics++ (FF++)~\cite{rossler2019faceforensics++}, DeepfakeDetection (DFD)~\cite{dfd}, Deepfake Detection Challenge (DFDC)~\cite{dolhansky2020deepfake}, and CelebDF (CDF)~\cite{li2020celeb}.
Note that there are three versions of FF++ in terms of compression level, \textit{i.e.,} raw, lightly compressed (c23), and heavily compressed (c40). Following previous face forgery detection works~\cite{li2020face,chen2022ost}, the FF++ (c23) is adopted.

\subsection{Implementation Details}
\label{ssec:Implementation Details}
We set the maximum size $N_m$ of the memory bank is 1000, the number $N_t$ of trainable transform layers to 5, and the dimension $d_t$ of the devised features to $d_t = d / 2$. The confidence threshold $Conf$ is set to 0.7, the number of nearby features $N_f$ is 16, the hyperparameter in the overall loss function $\alpha$ is 0.1. We use Adam~\cite{kingma2014adam} optimizer with a learning rate of 0.00001 and the updated step $K_s$ is 1. We implement our method in PyToch and conduct all experiments on a single NVIDIA RTX 4090 GPU with a test batch size of 32.

\begin{table}[tb!]
  \centering
  \caption{
  Comparison with state-of-the-art methods on CDF, DFD, and DFDC datasets using the AUC metric.
  All methods are trained on FF++ (c23).
  The best results are in bold and the second is underlined.
  }
  
  \scalebox{1.0}{

  \begin{tabular}{c|c|c|c|c|c} \toprule
    Detector & Publication & CDF & DFD & DFDC & Avg.\\ 
    \midrule
    Face X-ray~\cite{li2020face} & CVPR'20 & 0.679 & 0.766 & 0.633 & 0.693\\
    FFD~\cite{dang2020detection} & CVPR'20 & 0.744 & 0.802 & 0.703 & 0.750 \\
    F3Net~\cite{qian2020thinking} & ECCV'20 & 0.735 & 0.798 & 0.702 & 0.745 \\
    SPSL~\cite{liu2021spatial} & CVPR'21 & 0.765 & 0.812 & 0.704 & 0.760 \\
    Recce~\cite{cao2022end} & CVPR'22 & 0.732 & 0.812 & 0.713 & 0.752 \\
    UCF~\cite{yan2023ucf} & ICCV'23 & 0.753 & 0.807 & 0.719 & 0.760 \\
    LSDA~\cite{yan2024transcending} & CVPR'24 & 0.830 & \underline{0.880} & 0.736 & 0.815 \\
    CLIP~\cite{radford2021learning} & ICML'21 & \underline{0.846} & 0.849 & \underline{0.782} & \underline{0.826} \\
    \midrule
    CLIP + Ours & - & \textbf{0.941} & \textbf{0.937} & \textbf{0.862} & \textbf{0.913}\\
    \bottomrule
  \end{tabular}
  }

  \label{tab:cmp_sota}
  \vspace{-1pt}
\end{table}
\begin{table}[tb!]
  \centering
  \caption{
  Comparison with face forgery detection using test-time adaptation approach.
  They are trained on FF++ (c23) and tested on the CDF and DFDC datasets. For the metrics used, $\uparrow$ indicates that higher values are preferable, while $\downarrow$ signifies the opposite.
  }
  \scalebox{0.88}{
  \begin{tabular}{c|c|c|c|c|c|c} \toprule
    \multirow{2}{*}{Method} & \multicolumn{3}{c}{CDF} & \multicolumn{3}{c}{DFDC}\\ 
    \cmidrule(r){2-4} \cmidrule(r){5-7}  & AUC $\uparrow$ & ACC $\uparrow$ & ERR $\downarrow$  & AUC $\uparrow$ &ACC $\uparrow$  & ERR $\downarrow$ \\
    \midrule
    Xception~\cite{rossler2019faceforensics++} & 0.734 & 0.644 & 0.339 & 0.703 & 0.642 & 0.357\\
    OST~\cite{chen2022ost}& 0.748 & 0.673 & 0.312 & 0.833 & \textbf{0.714} & 0.250 \\
    \midrule
    Xception + Ours & \textbf{0.883} & \textbf{0.706} & \textbf{0.194} & \textbf{0.843} & 0.666 & \textbf{0.198}\\
    \bottomrule
  \end{tabular}}
  
  \label{tab:cmp_tta}
  \vspace{-1pt}
\end{table}
\subsection{Generalizability Comparisons}
\label{ssec:Results}
{\bf Comparison with state-of-the-art methods.}
To assess the generalization capacity of our method, we utilize CLIP~\cite{radford2021learning} as our base detector and compare it with several state-of-the-art face forgery detectors using the Area Under Curve (AUC) metric. These models are trained on FF++ (c23) and evaluated on other benchmark datasets.
As shown in TABLE~\ref{tab:cmp_sota}, our method significantly enhances the performance of the base detector (around 9\% on average) and achieves the overall best performance. While other detectors can generalize to some extent, they struggle with biases during evaluation and fall significantly short compared to our method. For example, our approach provides an average improvement of approximately 10\% over LSDA~\cite{yan2024transcending}. These results clearly highlight the advantage of our proposed test-time adaptation method.

{\bf Comparison with test-time adaptation-based methods in face forgery detection.}
For test-time adaptation-based methods in face forgery detection, we identify one prior work, OST~\cite{chen2022ost}, that shares similarities with our approach. Their framework aims to build an auxiliary task to facilitate test-time training. To ensure a fair comparison, we follow their experimental settings and utilize the Xception~\cite{rossler2019faceforensics++} as the base detector. In addition to the AUC metric, we also evaluate performance using accuracy (ACC) and Equal Error Rate (EER).
As reported in TABLE~\ref{tab:cmp_tta}, we observe that OST improves upon the baseline Xception, highlighting the importance of test-time adaptation. Additionally, our method outperforms OST on all testing datasets. This may be attributed to the fact that OST does not fully exploit the similarity within test data, which provides valuable information for detection.

\begin{table}[tb!]
  \centering
  \caption{
  Ablation studies on different test-time adaptation strategies. The baseline detector is CLIP trained on FF++(c23) and the metric is AUC.
  }
  \begin{tabular}{c|c|c|c|c} \toprule
    Detector & CDF & DFD & DFDC & Avg.\\ 
    \midrule
    Baseline  & 0.846 & 0.849 & 0.782 & 0.826 \\
    PL~\cite{lee2013pseudo} & 0.740 & 0.503 & 0.815 & 0.686 \\
    T3A~\cite{iwasawa2021test} & 0.838 & 0.867 & 0.741 & 0.815 \\
    TAST~\cite{jang2023test} & 0.917 & 0.912 & 0.840 & 0.890 \\
    Ours  & \textbf{0.941} & \textbf{0.937} & \textbf{0.862} & \textbf{0.913}\\
    \bottomrule
  \end{tabular}

  \label{tab:ablation_tta}
  \vspace{-1pt}
\end{table}

\begin{table}
  \centering
  \caption{
  Ablation studies regarding the effectiveness of different components, LCPC for the learnable class prototypes-based classifier, NFC for the nearest feature calibrator.
  }
  \begin{tabular}{c|c|c|c|c|c|c} \toprule
    ID & LCPC & NFC & CDF & DFD & DFDC & Avg.\\ 
    \midrule
    1 & $\times$ & $\times$  & 0.846 & 0.849 & 0.782 & 0.826 \\
    2 &\checkmark & $\times$ & 0.927 & 0.876 & 0.852  & 0.885  \\
    3 &$\times$ & \checkmark & 0.856 & 0.875 & 0.747  & 0.826  \\

   4 &\checkmark & \checkmark & \textbf{0.941} & \textbf{0.937} & \textbf{0.862} & \textbf{0.913} \\    
    
    \bottomrule
  \end{tabular}

  \label{tab:ablation_components}
\end{table}
{\bf Comparison with other test-time adaptation strategies.}
We conducted ablation studies on other test-time adaptation methods. As shown in TABLE~\ref{tab:ablation_tta}, the results demonstrate that our online test-time adaptation method achieves the best adaptation performance, highlighting its effectiveness. In contrast, PL~\cite{lee2013pseudo} shows reduced performance, likely due to the instability caused by updating the overall detector with pseudo-label self-training. Meanwhile, The T3A~\cite{iwasawa2021test} also fails to improve the baseline performance, potentially because the original features become less effective in the prototype-based classifier due to various forgery clues. TAST~\cite{jang2023test} shows significant improvement by incorporating a learnable adaptation layer and nearest-neighbor information but does not leverage information from the classifier of the base detector. Our method inherits the merits of TAST but differs by incorporating the predictions from the base detector as a supervision signal and utilizing the nearest-neighbor calibration, leading to superior performance improvements.

\subsection{Ablation Study}
{\bf Effects of different components.}
We conducted ablation studies to evaluate the contributions of learnable class prototype-based classifier (LCPC) and nearest feature calibrator (NFC) in our method. Specifically, we examined the following variants: (1) baseline, (2) the LCPC without the NFC, (3) the NFC without the LCPC (excluding the trainable transform layers in the LCPC), and (4) our overall method.
As shown in TABLE~\ref{tab:ablation_components}, comparing (1) with (2) shows that incorporating LCPC improves the baseline performance, demonstrating the value of the trainable transform layers and the prototype-based classification.
Comparing (3) with (4), we observed a significant improvement in performance (around 8.7\% on average) when including the trainable transform layers, indicating that they are crucial for effectively representing various forgery clues. 
Additionally, comparisons of (2) with (4) reveal that NFC further improves the performance of LCPC, highlighting the benefits of nearest feature calibration. 

\begin{figure}[!t] 
\centering 
\includegraphics[width=0.5\textwidth]{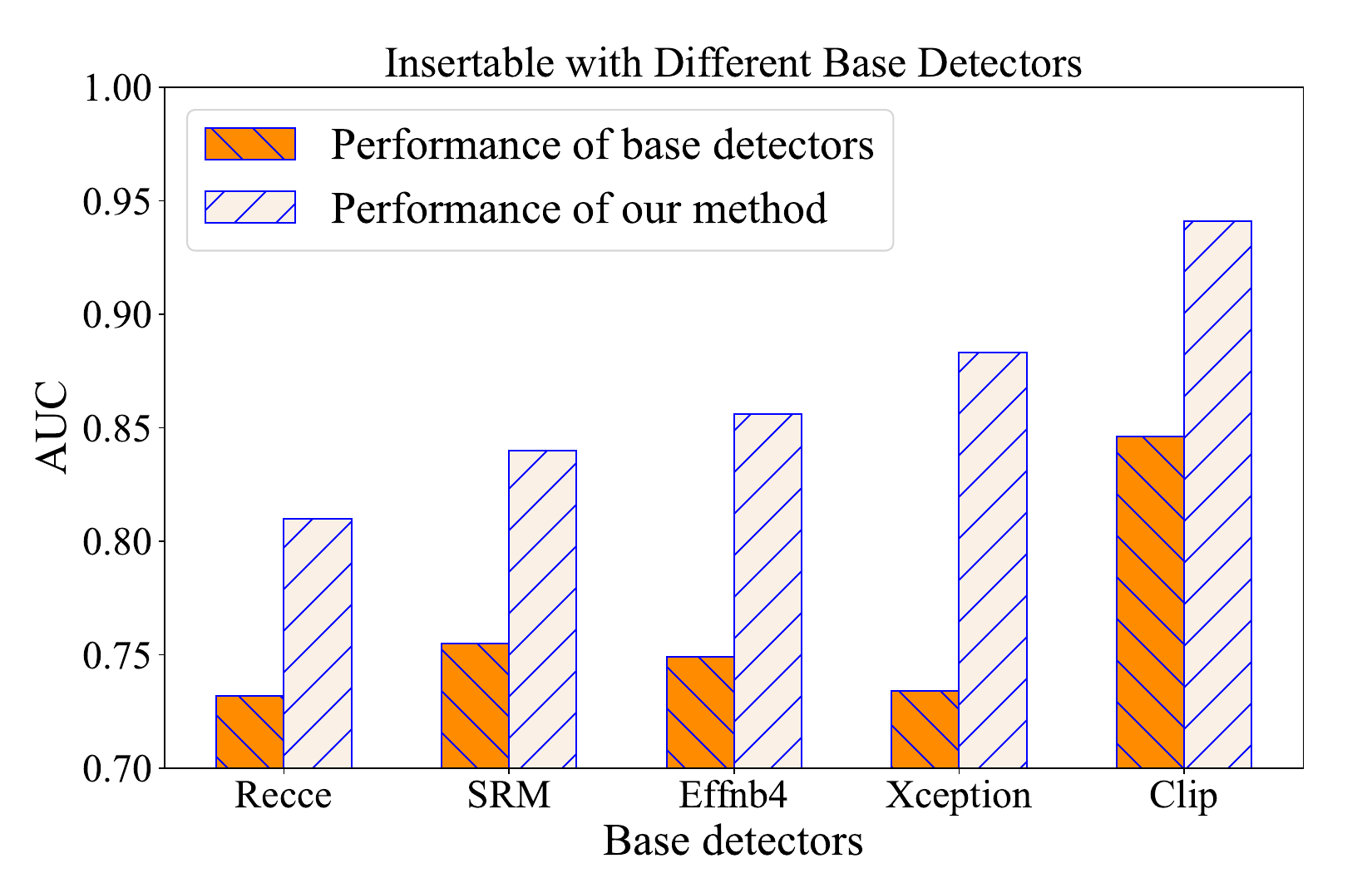} 
\caption{
Ablation studies on different base detectors. These detectors trained on FF++ (c23) and tested on CDF.
} 
\label{fig:insertable} 
\end{figure}

\begin{table}[tb!]
  \centering
  \caption{
  Performance of using different updated steps. We report AUC on the DFD dataset and time consummations (TC).
  }
  \scalebox{0.95}{
  \begin{tabular}{c|c|c|c} \toprule
    Updated Steps & 1 step & 2 step & 3 step  \\ 
    \midrule
    DFD & 0.937 & 0.941 & 0.945 \\
    TC(s) & 0.035s  & 0.063s & 0.086s  \\
    \bottomrule
  \end{tabular}
  }

  \label{tab:ablation_TC}
  \vspace{-1pt}
\end{table}

{\bf Effects of different update steps.}
To investigate whether increasing the number of update steps improves performance, we conduct ablation studies with varying update steps $K_s$ during evaluation. As shown in TABLE~\ref{tab:ablation_TC}, the accuracy improves with more update steps, while computational time increases proportionally. For a balance between accuracy and efficiency, a single update step provides satisfactory results.

{\bf Insertable with different base detectors.}
To demonstrate the compatibility of our adaptation module with a range of base detectors, we conducted ablation studies by combining our module with several detectors including Recce~\cite{cao2022end}, SRM~\cite{luo2021generalizing}, Effinb4~\cite{tan2019efficientnet}, Xception~\cite{rossler2019faceforensics++}, and CLIP~\cite{radford2021learning}. These detectors feature diverse architectures and training procedures. As shown in Fig.~\ref{fig:insertable}, our method enhances the performance of these detectors, showcasing the versatility of our module.

\section{CONCLUSION}
\label{sec:CONCLUSION}

In this paper, we introduce an insertable adaptation module designed to enhance performance of a trained face forgery detector during evaluation, without requiring modifications to the architecture or training process. Our method integrates LCPC and NFC, enabling the module to effectively process diverse forgery clues and improve prediction accuracy during testing. Experiments across multiple datasets show that our module achieves superior generalization compared to state-of-the-art methods. Moreover, it significantly enhances the performance of various detectors, highlighting its versatility and effectiveness.

\vfill\pagebreak
\bibliographystyle{IEEEtran}
\bibliography{IEEEabrv,reference_bak}

\vspace{12pt}

\end{document}